\documentclass[10pt,twocolumn,letterpaper]{article}

\usepackage{cvpr}
\usepackage{times}
\usepackage{epsfig}
\usepackage{graphicx}
\usepackage{amsmath}
\usepackage{amssymb}
\usepackage{amstext}
\usepackage{multirow}
\usepackage{booktabs}
\usepackage[breaklinks=true,bookmarks=false]{hyperref}

\cvprfinalcopy % *** Uncomment this line for the final submission

\def\cvprPaperID{****} % *** Enter the CVPR Paper ID here

\begin{document}

%%%%%%%%% TITLE
\title{Object Skeleton Extraction in Natural Images by Fusing Scale-associated Deep Side Outputs}

\author{Wei Shen$^1$, Kai Zhao$^1$, Yuan Jiang$^1$, Yan Wang$^2$, Zhijiang Zhang$^1$, Xiang Bai$^3$\\
$^1$ Key Laboratory of Specialty Fiber Optics and Optical Access Networks, Shanghai University\\
$^2$ Rapid-Rich Object Search Lab, Nanyang Technological University\\
$^3$ School of Electronic Information and Communications, Huazhong University of Science and Technology\\
{\tt\footnotesize
wei.shen@shu.edu.cn,zeakey@outlook.com,jy9387@outlook.com,wyanny.9@gmail.com,zjzhang@shu.edu.cn,xbai@hust.edu.cn}}

\maketitle
%\thispagestyle{empty}

%%%%%%%%% ABSTRACT
\begin{abstract}
Object skeleton is a useful cue for object detection, complementary to the object contour, as it provides a structural representation to describe the relationship among object parts. While object skeleton extraction in natural images is a very challenging problem, as it requires the extractor to be able to capture both local and global image context to determine the intrinsic scale of each skeleton pixel. Existing methods rely on per-pixel based multi-scale feature computation, which results in difficult modeling and high time consumption. In this paper, we present a fully convolutional network with multiple scale-associated side outputs to address this problem. By observing the relationship between the receptive field sizes of the sequential stages in the network and the skeleton scales they can capture, we introduce a scale-associated side output to each stage. We impose supervision to different stages by guiding the scale-associated side outputs toward groundtruth skeletons of different scales. The responses of the multiple scale-associated side outputs are then fused in a scale-specific way to localize skeleton pixels with multiple scales effectively. Our method achieves promising results on two skeleton extraction datasets, and significantly outperforms other competitors.
\end{abstract}

%%%%%%%%% BODY TEXT
\section{Introduction} \label{sec:intro}
\begin{figure}[!h]
\centering
\includegraphics[width=1.0\linewidth]{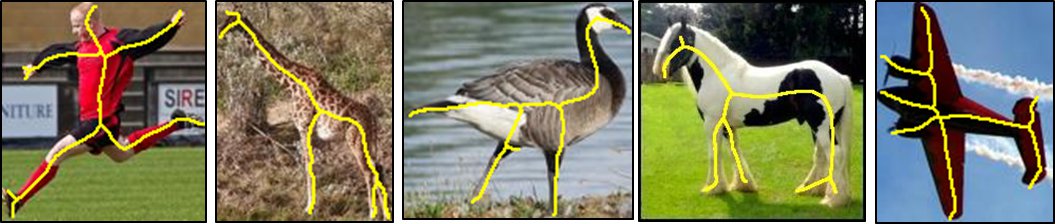}
\caption{Object skeleton extraction in natural images. The skeletons are in yellow.}\label{fig:sk_ex}
\end{figure}

In this paper, we investigate an interesting and nontrivial problem in computer vision, object skeleton extraction from natural images (Fig.~\ref{fig:sk_ex}). Here, the concept of ``object'' means a standalone thing with a well-defined boundary and center~\cite{Ref:AlexeDF10}, such as an animal, a human, and a plane, as opposed to amorphous background stuff, such as sky, grass, and mountain. Skeleton, also called \emph{symmetry axis}, is a useful structure-based object descriptor. Extracting object skeletons directly from natural images is of broad interests to many real applications including object recognition/detection~\cite{Ref:Bai09,Ref:TrinhK11}, text recognition~\cite{Ref:Zhang15}, road detection and blood vessel detection~\cite{Ref:SironiLF14}.

Skeleton extraction from pre-segmented images~\cite{Ref:Saha15} used to be a hot topic, which has been well studied and successfully applied to shape-based object matching and recognition~\cite{Ref:Siddiqi99,Ref:Sebastian04,Ref:Demirci06}. However, such methods have severe limitations when being applied to natural images, as segmentation from natural images is still an unsolved problem.

Skeleton extraction from natural images is a much more challenging problem. The main difficulties stem from three aspects: (1) Complexity of natural scenes: Natural scenes can be very cluttered. Amorphous background elements, such as fences, bricks and even the shadows of objects, exhibit somewhat self-symmetry, and thus are prone to cause distractions. (2) Diversity of objects: Objects in natural images may exhibit entirely different colors, textures, shapes and sizes. (3) Specificity of skeletons: local skeleton segments have a variety of patterns, such as straight lines, T-junctions and Y-junctions. In addition, a local skeleton segment naturally associates with a certain scale, determined by the thickness of its corresponding object part. However, it is unknown in natural images. We term this problem as unknown-scale problem in skeleton extraction.

A number of works have been proposed to study this problem in the past decade. Broadly speaking, they can be categorized into two groups: (1) Traditional image processing methods~\cite{Ref:Yu04,Ref:Jang01,Ref:Lindeberg98,Ref:ZhangC07}, which compute skeletons from a gradient intensity map according to some geometric constraints between edges and skeletons. Due to the lack of
object prior, these methods can not handle the images with complex scenes; (2) Recent learning based methods~\cite{Ref:Tsogkas12,Ref:Levinshtein09,Ref:Lee13,Ref:SironiLF14,Ref:Widynski14}, which learn a per-pixel classification or segment-linking model based on elaborately hand-designed features computed at multi-scales for skeleton extraction. Limited by the ability of traditional learning models and hand-designed features, these methods fail to extract the skeletons of objects with complex structures and cluttered interior textures. In addition, such per-pixel/segment models are usually quite time consuming for prediction. Consequently, there still remains obvious gap between these skeleton extraction methods and human perception, in both performance and speed. Skeleton extraction has its unique aspect by looking into both local and global image context, which requires much more powerful models in both multi-scale feature learning and classifier learning, since the visual complexity increases exponentially with the size of the context field.

To tackle the obstacles mentioned above, we develop a holistically-nested network with multiple scale-associated side outputs for skeleton extraction. The holistically-nested network~\cite{Ref:Xie15} is a deep fully convolutional network (FCN)~\cite{Ref:LongSD15}, which enables holistic image training and prediction for per-pixel tasks. Here, we connect a scale-associated side output to each convolutional layer in the holistically-nested network to address the unknown-scale problem in skeleton extraction.

\begin{figure}[!t]
\centering
\includegraphics[width=1.0\linewidth]{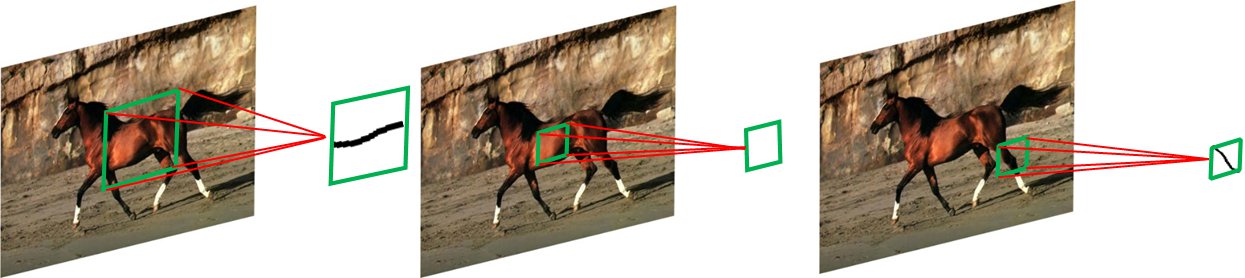}
\caption{Using filters (the green squares on images) of multiple sizes for skeleton extraction. Only when the size of the filter is larger than the scale of current skeleton part can the filter capture enough context feature to detect it.}\label{fig:resp}
\end{figure}

Referring to Fig.~\ref{fig:resp}, imagine that we are using multiple filters with different sizes (such as the convolutional kernels in convolutional networks) to detect a skeleton pixel with a certain scale; then only the filters with the sizes larger than the scale will have responses on it, and others will not. Note that the sequential convolutional layers in a holistically-nested network can be treated as the filters with increasing sizes (the receptive field sizes on the original image of each convolutional layer are increasing from shallow to deep). So each convolutional layer is only able to capture the features of the skeleton pixels with scales less than its receptive field size. The sequential increasing receptive field sizes provide a principle to quantize the skeleton scale space. With these observations, we propose to impose supervision to each side output, optimizing it towards a scale-associated groundtruth skeleton map. More specifically, each skeleton pixel in it is labeled by a quantized scale value and only the skeleton pixels whose scales are smaller than the receptive filed size of the side output are reserved. Thus, each side output is associated with some certain scales and able to give a certain number of scale-specific skeleton score maps (the score map for one specified quantized scale value) when predicting.

The final predicted skeleton map can be obtained by fusing these scale-associated side outputs. A straightforward fusion method is to average them. However, a skeleton pixel with larger scale probably has a stronger response on a deeper side output, and a weaker response on a shallower side output; a skeleton pixel with smaller scale may have strong responses on both of the two side outputs. By considering this phenomenon, for each quantized scale value, we propose to use a scale-specific weight layer to fuse the corresponding scale-specific skeleton score map provided by each side output.

In summary, the core contribution of this paper is the proposal of the scale-associated side output layer, which enables both target learning and fusion in a scale-associated way. Therefore, our holistically-nested network is able to localize skeleton pixels with multiple scales.

\begin{figure}[!h]
\centering
\includegraphics[width=1.0\linewidth]{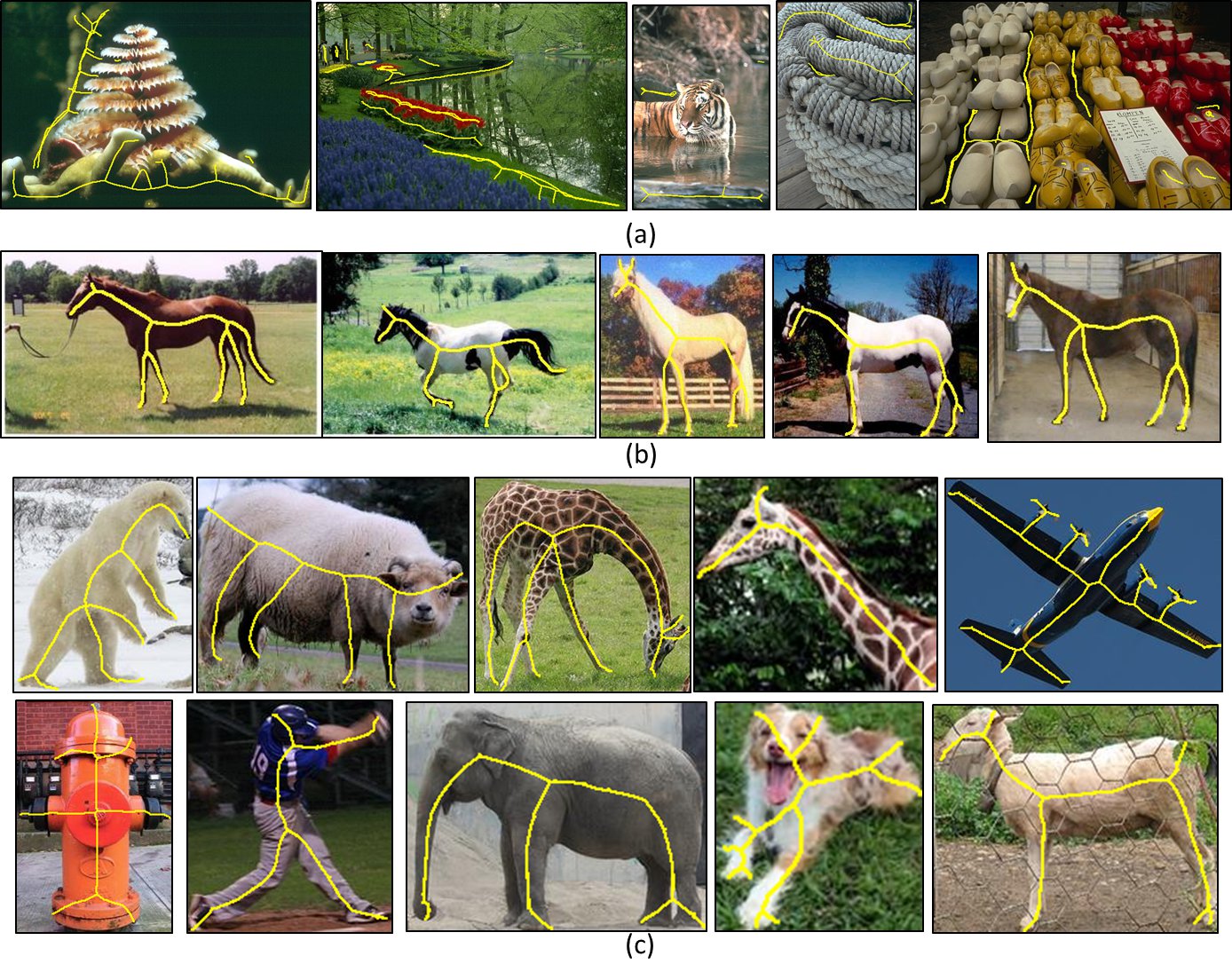}
\caption{Some samples from three datasets. (a) The SYMMAX300 dataset~\cite{Ref:Tsogkas12}. (b) The WH-SYMMAX dataset~\cite{Ref:Shen16}. (c) Our new dataset, the SK506 dataset. The groundtruths for skeleton or local reflection symmetry are in yellow.}\label{fig:dataset}
\end{figure}

To the best of our knowledge, there are only two datasets related to our task. One is the SYMMAX300 dataset~\cite{Ref:Tsogkas12}, which is converted from the well-known Berkeley Segmentation Benchmark (BSDS300)~\cite{Ref:Martin01}. However, this dataset is used for local reflection symmetry detection. Local reflection symmetry~\cite{Ref:liu09,Ref:LeeL12} is a kind of low-level feature of image, regardless of the concept of ``object''. Some samples in this dataset are shown in Fig.~\ref{fig:dataset}(a). Note that, a large number of symmetries occur in non-object parts. Generally, object skeleton is a subset of local reflection symmetry. The other one is the WH-SYMMAX dataset~\cite{Ref:Shen16}, which is converted from the Weizmann Horse dataset~\cite{Ref:Borenstein02}. This dataset is suitable to verify object skeleton extraction methods; however, as shown in Fig.~\ref{fig:dataset}(b) the limitation is that only one object category, the horse, is contained in it. To evaluate skeleton extraction methods, we construct a new dataset, named SK506\footnote{http://wei-shen.weebly.com/uploads/2/3/8/2/23825939/sk506.zip}. There are 506 natural images in this dataset, which are selected from the recent published MS COCO dataset~\cite{Ref:Chen15}. The objects in these 506 images belong to a variety of categories, including humans, animals, such as birds, dogs and giraffes, and artificialities, such as planes and hydrants. We apply a skeletonization method~\cite{Ref:Bai07} to the provided human-annotated foreground segmentation maps of the selected images to generate the groundtruth skeleton maps. Some samples of the SK506 dataset are shown in Fig.~\ref{fig:dataset}(c). We evaluate several skeleton extraction methods as well as symmetry detection methods on both SK506 and WH-SYMMAX. The experimental results demonstrate that the proposed method significantly outperforms others.
%Both edge and skeleton are line features of images, and they are indeed related to each other; however, as a local skeleton segment naturally associates a certain scale while a local edge segment does not, methods for skeleton extraction should be more complicated.  With these observations, we propose to impose supervision to each side output, guiding it toward a scale-associated skeleton groudtruth map, in which only the skeleton pixels whose scales are larger than the receptive filed size of the side output are reserved.
%-------------------------------------------------------------------------
\section{Related Works}
Object skeleton extraction has been paid much attention in previous decades. However, most works in the early stage~\cite{Ref:Saha15,Ref:Bai07} only focus on skeleton extraction from pre-segmented images. As these works have a strict assumption that object silhouettes are required to be available, they cannot be applied in our task.

Some pioneers try to extract skeletons from the gradient intensity maps computed on natural images. The gradient intensity map is generally obtained by applying directional derivative operators to a gray-scale image smoothed by a Gaussian kernel. For instance, in~\cite{Ref:Lindeberg98}, the author provides an automatic mechanism to determine the best size of the Gaussian kernel used for gradient computation, and he propose to detect skeletons as the pixels for which the gradient intensity assumes a local maximum (minimum) in the direction of the main principal curvature. Jang and Hong~\cite{Ref:Jang01} extract the skeleton from the pseudo-distance map which is obtained by iteratively minimizing an object function defined on the gradient intensity map. Yu and Bajaj~\cite{Ref:Yu04} propose to trace the ridges of the skeleton intensity map calculated from the diffused vector field of the gradient intensity map, which can remove the undesirable biased skeletons. Due to the lack of object prior, these methods are only able to handle the images with simple scenes.

Recent learning based skeleton extraction methods are more suitable to deal with the scene complexity problem in natural images. One type of them formulates skeleton extraction to be a per-pixel classification problem. Tsogkas and Kokkinos~\cite{Ref:Tsogkas12} compute the hand-designed features of multi-scale and multi-orientation at each pixel, and employ the multiple instance learning framework to determine whether it is symmetry\footnote{Although symmetry detection is not the same problem as skeleton extraction, we also compare the methods for it with ours, as skeleton can be considered a subset of symmetry.} or not. Shen \emph{et al.}~\cite{Ref:Shen16} then improve their method by training MIL models on automatically learned scale- and orientation-related subspaces. Sironi \emph{et al.}~\cite{Ref:SironiLF14} transform the per-pixel classification problem to a regression one to achieve accurate skeleton localization, which learns the distance to the closest skeleton segment in scale-space.  Alternatively, another type of learning based methods aim to learn the similarity between local skeleton segments (represented by superpixel~\cite{Ref:Levinshtein09,Ref:Lee13} or spine model~\cite{Ref:Widynski14}), and link them by hierarchical clustering~\cite{Ref:Levinshtein09}, dynamic programming~\cite{Ref:Lee13} or particle filter~\cite{Ref:Widynski14}. Due to the limited power of the hand-designed features and traditional learning models, these methods are intractable to detect the skeleton pixels with large scales, as much more context information is needed to be handled.

Our method is inspired by~\cite{Ref:Xie15}, which develops a holistically-nested network for edge detection (HED). Edge detection does not face the unknown-scale problem. Using a local filter to detect an edge pixel, no matter what the size of the filter is, will have responses, either stronger or weaker. So summing up the multi-scale detection responses, which is adopted in the fusion layer in HED, is able to improve the performance of edge detection~\cite{Ref:Ren08,Ref:DollarZ15,Ref:Shen15}, while bringing noises across the scales for skeleton extraction. There are two main differences between HED and our method. 1. We supervise the side outputs of the network with different scale-associated groundtruths, while the groundtruths in HED are the same. 2. We use different scale-specific weight layers to fuse the corresponding scale-specific skeleton score maps provided by the side outputs, while the side outputs are fused by a single weight layer in HED. Such two changes utilize multi stages in a network to explicitly detect the unknown scale, which HED is unable to handle with. With the extra supervision added to each layer, our method is able to provide a more informative result, i.e., the predicted scale for each skeleton pixel, which is useful for other potential applications, such as part segmentation and object proposal detection (we will show this in Sec.~\ref{sec:part_seg} and Sec.~\ref{sec:detection}). While the result of HED cannot be applied to such applications.

\section{Methodology}
In this section, we describe our methods for object skeleton extraction. First, we introduce the architecture of our holistically-nested network. Then, we discuss how to optimize and fuse the multiple scale-associated side outputs in the network for skeleton extraction.
\subsection{Network Architecture}
\begin{figure}[!h]
\centering
\includegraphics[width=1.0\linewidth]{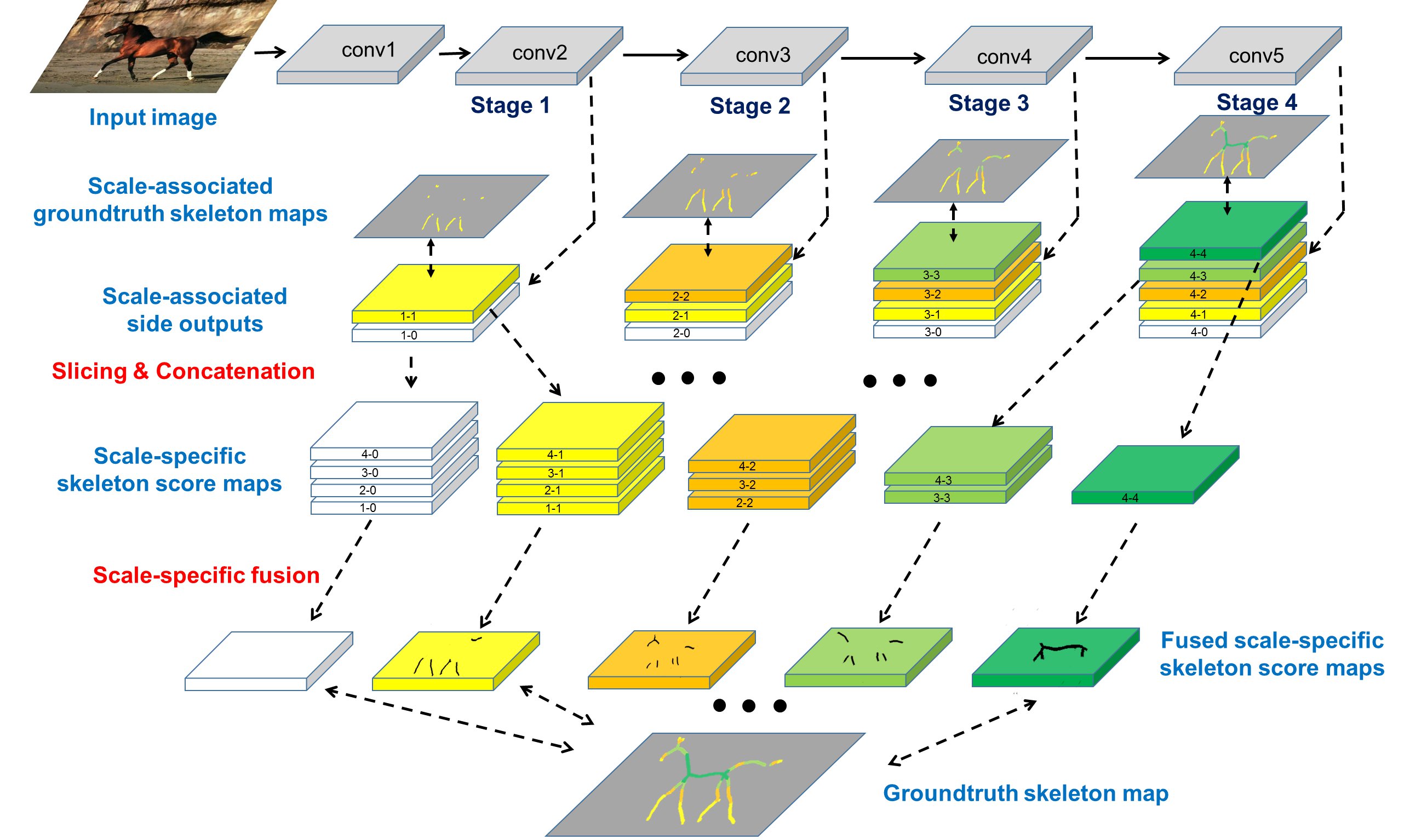}
\caption{The proposed network architecture for skeleton extraction, which is converted from VGG 16-layer net~\cite{Ref:Simonyan14}. It has 4 stages with additional scale-associated side output layers connected to the convolutional layers. Each scale-associated side output is guided by a scale-associated groundtruth skeleton map (The skeleton pixels with different quantized scales are in different colors.). Each scale-associated side output layer provides a certain number of scale-specific skeleton score maps (identified by stage number-quantized scale value pairs). The score maps of the same scales from different stages will be sliced and concatenated. Five scale-specific weighted-fusion layers are added to automatically fuse outputs from multiple stages.
}\label{fig:network}
\end{figure}

The recent work~\cite{Ref:Agrawal14} has demonstrated that fine-tuning well pre-trained deep neural networks is an efficient way to obtain a good performance on a new task. Therefore, we basically adopt the network architecture used in~\cite{Ref:Xie15}, which is converted from VGG 16-layer net~\cite{Ref:Simonyan14} by adding additional side output layers and replacing fully-connected layers by fully-convolutional layers with $1\times1$ kernel size. Each fully-convolutional layer is then connected to an upsampling layer to ensure that the outputs of all the stages are with the same size. Here, we make several modifications for our task skeleton extraction: (a) we connect the proposed scale-associated side output layer to the last convolutional layer in each stage except for the first one, respectively conv2\_2, conv3\_3, conv4\_3, conv5\_3. The receptive field sizes of the scale-associated side output layers are 14, 40, 92, 196, respectively. The reason why we omit the first stage is that the receptive field size of the last convolutional layer in it is too small (only 5) to capture any skeleton features. There are few skeleton pixels with scales less than such a small receptive field size. (b) Each scale-associated side output layer provides a certain number of scale-specific skeleton score maps. Each scale-associated side output layer is connected to a slice layer to obtain the skeleton score map for each scale. Then from all the scale-associated side output layers, we use a scale-specific weight layer to fuse the skeleton score maps for this scale. Such a scale-specific weight layer can be achieved by a fully-convolutional layer with $1\times1$ kernel size. In this way, the skeleton score maps for different scales are fused by different weight layers. The fused skeleton score maps for each scale are concatenated together to form the final predicted skeleton map. To sum up, our holistically-nested network architecture has 4 stages with additional scale-associated side output layers, with strides 2, 4, 8 and 16, respectively, and with different receptive field sizes; it also has 5 additional weight layers to fuse the side outputs. An illustration for the network architecture is shown in Fig.~\ref{fig:network}.

\subsection{Skeleton Extraction by Fusing Scale-associated Side Outputs}
Skeleton extraction can be formulated as a per-pixel classification problem. Given a raw input image $X=\{x_j,j=1,\ldots,|X|\}$, our goal is to predict its skeleton map $\hat{Y}=\{\hat{y}_j,j=1,\ldots,|X|\}$, where $\hat{y}_j\in\{0,1\}$ denotes the predicted label for each pixel $x_j$, i.e., if $x_j$ is predicted as a skeleton pixel, $\hat{y}_j=1$; otherwise, $\hat{y}_j=0$. Next, we describe how to learn and fuse the scale-associated side outputs in the training phase as well as how to utilize the learned network in the testing phase, respectively.
\subsubsection{Training Phase} \label{sec:training}

We are given a training dataset denoted by $\mathcal {S}=\{(X^{(n)},Y^{(n)}),n=1,\ldots,N\}$, where $X^{(n)}=\{x^{(n)}_j,j=1,\ldots,|X^{(n)}|\}$ is a raw input image and $Y^{(n)}=\{y^{(n)}_j,j=1,\ldots,|X^{(n)}|\}$ ($y^{(n)}_j\in\{0,1\}$) is its corresponding groundtruth skeleton map. First, we describe how to compute a quantized skeleton scale map for each training image, which will be used for guiding the network training.
\paragraph{Skeleton scale quantization.}
According to the definition of skeleton~\cite{Ref:Blum67}, we define the scale of each skeleton pixel as the diameter of the maximal disk centered at it, which can be obtained when computing the groundtruth skeleton map from the pre-segmented image. By defining the scale of each non-skeleton pixel to be zero, we build a scale map $S^{(n)}=\{s^{(n)}_j,j=1,\ldots,|X^{(n)}|\}$ for each $Y^{(n)}$ and we have $y^{(n)}_j=\mathbf{1}(s^{(n)}_j>0)$, where $\mathbf{1}(\cdot)$ is an indicator function. As we consider each image holistically, we drop the superscript $n$ in our notation. We aim to learn a holistically-nested network with multiple stages of a convolutional layer linked with a scale-associated side output layer. Assume that there are $M$ such stages in our network, in which the receptive field sizes of the convolutional layers increase sequentially. Let $(r_i;i=1,\ldots,M)$ be the sequence of the receptive field sizes. Recall that only when the receptive field size is larger than the scale of a skeleton pixel can the convolutional layer capture the features of it. Thus, the scale of a skeleton pixel can be quantized into a discrete value, to indicate which stages in the network are able to detect this skeleton pixel. (Here, we assume that $r_M$ is sufficiently large for capturing the features of the skeleton pixels with the maximum scale). The quantized value $z$ of a scale $s$ is computed by

\begin{equation}
z=\left\{
\begin{aligned}
&\arg\min_{i=1,\ldots,M} i, \; \text{s.t.} \; {r_i}>\lambda{s}   &\text{if} \; s>0\\
&0 &\text{if} \; s=0 \\
\end{aligned}
\right.,
\end{equation}
where $\lambda>1$ is a factor to ensure that the receptive field sizes are sufficiently large for feature computation. (We set $\lambda=1.2$ in our experiments.)
Now, for an image $X$, we can build a quantized scale value map $Z=\{z_j,j=1,\ldots,|X|\}\}$($z_j\in\{0,1,\ldots,M\}$).
\paragraph{Scale-associated side output learning.}
The groundtruth skeleton map $Y$ can be trivially converted from $Z$: $Y=\mathbf{1}(Z>0)$, but not vice versa. So we would like to guild the network training by $Z$ instead of $Y$, as more supervision can be included. This actually convert a binary classification problem to a multi-class classification, where each class corresponds a quantized scale. Towards this end, each side output layer in our network is associated with a softmax regression classifier. While according to the above discussions, one stage in our network is only able to detect the skeleton pixels with scales less than its corresponding receptive field size. Therefore, the side output is scale-associated. For the $i$-th side output, we guild it to a scale-associated groundtruth skeleton map: $Z^{(i)}=Z\circ{\mathbf{1}(Z\leq{i})}$, where $\circ$ is an element-wise product operator. Let $K^{(i)}$ denote the maximum value in $Z^{(i)}$, i.e., $K^{(i)}=i$, then we have $Z^{(i)}=\{z^{(i)}_j,j=1,\ldots,|X|\},z^{(i)}_j\in\{0,1,\ldots,K^{(i)}\}$. Let $\ell_{s}^{(i)}(\mathbf{W},\mathbf{\Phi}^{(i)})$ denote the loss function for this scale-associated side output, where $\mathbf{W}$ and $\mathbf{\Phi}^{(i)}$ are the layer parameters of the network and the parameters of the classifier of this stage. As our network enables holistic image training, the loss function is computed over all pixels in the training image $X$ and the scale-associated groundtruth skeleton map $Z^{(i)}$. Generally, the distribution of skeleton pixels with different scales and non-skeleton pixels is biased in an image. Therefore, we define a weighted softmax loss function to balance the loss between these multiple classes:
\begin{equation}
\begin{aligned}
&\ell_{s}^{(i)}(\mathbf{W},\mathbf{\Phi}^{(i)})=  \\
&-\frac{1}{|X|}\sum_{j=1}^{|X|}\sum_{k=0}^{K^{(i)}}\beta_k^{(i)}\mathbf{1}(z^{(i)}_j=k)\log\text{Pr}(z^{(i)}_j=k|X;\mathbf{W},\mathbf{\Phi}^{(i)}),
\end{aligned}
\end{equation}
where $\beta_k^{(i)}$ is the loss weight for the $k$-th class and $\text{Pr}(z^{(i)}_j=k|X;\mathbf{W},\mathbf{\Phi}^{(i)})\in[0,1]$ is the predicted score given by the classifier for how likely the quantized scale of $x_j$ is $k$. $\mathcal {N}(\cdot)$ denotes the number of non-zero elements in a set, then $\beta_k$ can be computed by
\begin{equation} \label{eqn:beta}
\beta_k^{(i)}=\frac{\frac{1}{\mathcal {N}(\mathbf{1}(Z_i==k))}}{\sum_{k=0}^{K^{(i)}}\frac{1}{\mathcal {N}(\mathbf{1}(Z_i==k))}}.
\end{equation}
Let $a^{(i)}_{jk}$ be the activation of the $i$-th side output associated with the quantized scale $k$ for the input $x_j$, then we use the softmax function~\cite{Ref:Bishop06} $\sigma(\cdot)$ to compute
\begin{equation}
\text{Pr}(z^{(i)}_j=k|X;\mathbf{W},\mathbf{\Phi}^{(i)})=\sigma(a^{(i)}_{jk})=\frac{\exp(a^{(i)}_{jk})}{\sum_{k=0}^{K^{(i)}}\exp(a^{(i)}_{jk})}.
\end{equation}
One can show that the partial derivation of $\ell_{s}^{(i)}(\mathbf{W},\mathbf{\Phi}^{(i)})$ w.r.t. $a^{(i)}_{jl}$ ($l\in\{0,1,\ldots,K^{(i)}\}$) can be obtained by
\begin{equation}
\begin{aligned}
&\frac{\partial{\ell_{s}^{(i)}(\mathbf{W},\mathbf{\Phi}^{(i)})}}{\partial{a^{(i)}_{jl}}}=-\frac{1}{m}\bigg(\beta_{l}^{(i)}\mathbf{1}(z^{(i)}_j=l)-\\
&\sum_{k=0}^{K^{(i)}}\beta_{k}^{(i)}\mathbf{1}(z^{(i)}_j=k)\text{Pr}(z^{(i)}_j=l|X;\mathbf{W},\mathbf{\Phi}^{(i)})\bigg).
\end{aligned}
\end{equation}
$\mathbf{\Phi}=(\mathbf{\Phi}^{(i)};i=1,\ldots,M)$ denotes the parameters of the classifiers in all the stages, then the loss function for all the side outputs is simply obtained by
\begin{equation}
\mathcal {L}_s(\mathbf{W},\mathbf{\Phi})=\sum_{i=1}^M\ell_{s}^{(i)}(\mathbf{W},\mathbf{\Phi}^{(i)}).
\end{equation}
\paragraph{Multiple scale-associated side outputs fusion.}
For an input pixel $x_j$, each scale-associated side output provides a predicted score $\text{Pr}(z^{(i)}_j=k|X;\mathbf{W},\mathbf{\Phi}^{(i)})$ (if $k{\leq}K^{(i)}$) for representing how likely its quantized scale is $k$. We can obtain a fused score $f_{jk}$ by simply summing them with weights $\mathbf{a}_k=({a}^{(i)}_k;i=\max(k,1),\ldots,M)$:
\begin{equation}
\begin{aligned}
f_{jk}=\sum^M_{i=\max(k,1)}{a}^{(i)}_k\text{Pr}(z^{(i)}_j=k|X;\mathbf{W},\mathbf{\Phi}^{(i)}),\\\text{s.t.}\;\sum_{i=\max(k,1)}^M{a}^{(i)}_k=1.
\end{aligned}
\end{equation}
We can understand the above fusion by this way: each scale-associated side output provides a certain number of scale-specific predicted skeleton score maps, and we utilize $M+1$ scale-specific weight layers: $\mathbf{A}=(\mathbf{a}_k;k=0,\ldots,M)$ to fuse them. Similarly, we can define a fusion loss function by
\begin{equation}
\begin{aligned}
&\mathcal {L}_f(\mathbf{W},\mathbf{\Phi},\mathbf{A})=  \\
&-\frac{1}{|X|}\sum_{j=1}^{|X|}\sum_{k=0}^{M}\beta_k\mathbf{1}(z_j=k)\log\text{Pr}(z_j=k|X;\mathbf{W},\mathbf{\Phi},\mathbf{a}_k),
\end{aligned}
\end{equation}
where $\beta_k$ is defined by the same way in Eqn.~\ref{eqn:beta} and $\text{Pr}(z_j=k|X;\mathbf{W},\mathbf{\Phi},\mathbf{w}_k)=\sigma(f_{jk})$.

Finally, we can obtain the optimal parameters by
\begin{equation}
(\mathbf{W},\mathbf{\Phi},\mathbf{A})\ast =\arg\min(\mathcal {L}_s(\mathbf{W},\mathbf{\Phi})+\mathcal {L}_f(\mathbf{W},\mathbf{\Phi},\mathbf{A})).
\end{equation}
\subsubsection{Testing Phase}
Given a testing image $X=\{x_j,j=1,\ldots,|X|\}$, with the learned network $(\mathbf{W},\mathbf{\Phi},\mathbf{A})\ast$, its predicted skeleton map $\hat{Y}=\{\hat{y}_j,j=1,\ldots,|X|\}$ is obtained by
\begin{equation} \label{eqn:predict}
\hat{y}_j = 1 - \text{Pr}(z_j=0|X;\mathbf{W}\ast,\mathbf{\Phi}\ast,\mathbf{a}_0\ast).
\end{equation}
Recall that $z_j=0$ and $z_j>0$ mean that $x_j$ is a non-skeleton/skeleton pixel, respectively. We refer to our method as FSDS, for fusing scale-associated deep side outputs.
\subsection{Understanding of the Proposed Method}
\begin{figure}[!h]
\centering
\includegraphics[width=1.0\linewidth]{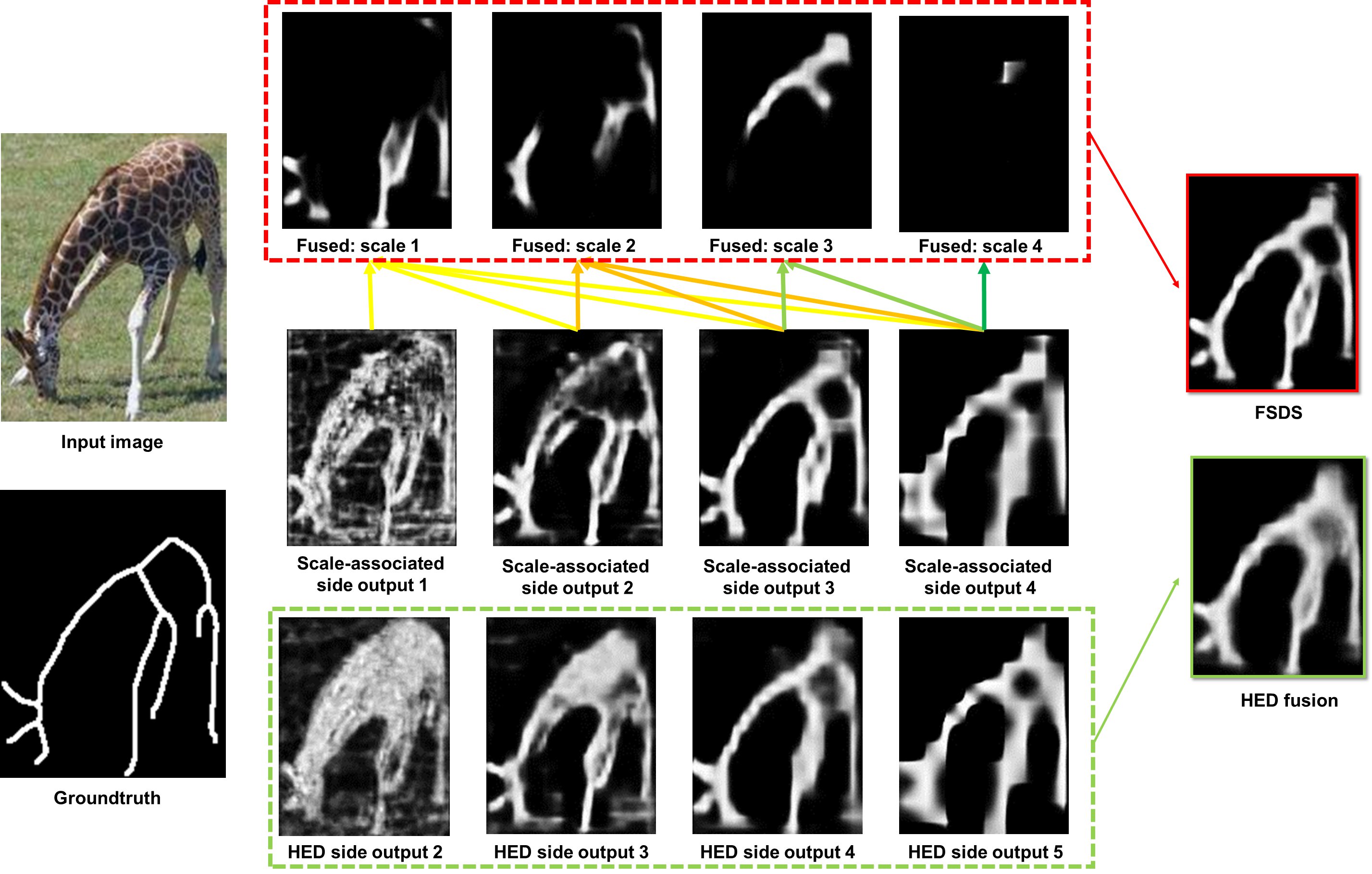}
\caption{The comparison between the intermediate results of FSDS and HED. We can observe that the former are able to differentiate skeleton pixels with different scales, while the latter cannot.
}\label{fig:inter}
\end{figure}

To understand our method more deeply, we illustrate the intermediate results of our method and compare with those of HED in Fig.~\ref{fig:inter}. The response of each scale-associated side output can be obtained by the similar way of Eqn.~\ref{eqn:predict}. We compare the response of each scale-associated side output to the corresponding one in HED (The side output 1 in HED is connected to conv1\_2, while ours start from conv2\_2.). With the extra scale-associated supervision, the responses of our side outputs are indeed related to scale. For example, the first one fires on the structure with small scales, such as the legs, the interior textures and the object boundaries; while in the second one, the skeleton parts of the head and neck are clear and meanwhile the noises on small scale structure are suppressed. In addition, we perform scale-specific fusion, by which each fused scale-specific skeleton score map indeed corresponds to one scale (See the first three response maps corresponding to legs, neck and torso respectively). The side outputs in HED are not able to differentiate skeleton pixels with different scales. Consequently, the first two respond on the whole body, which bring noises to the final fusion one.
%Similarly, the skeleton map predicted by each scale-associated side output layer $\hat{Y}^{(i)}=\{\hat{y}^{(i)}_j,j=1,\ldots,|X|\}$ is obtained by
%\begin{equation}
%\hat{y}^{(i)}_j = 1 - \text{Pr}(z^{(i)}_j=0|X;\mathbf{W}\ast,\mathbf{\Phi}^{(i)}\ast).
%\end{equation}
\section{Experimental Results}
In this section we discuss the implementation details and compare the performance of our skeleton extraction methods with competitors.
\subsection{Implementation Details}
Our architecture is built on the public available implementation of FCN~\cite{Ref:LongSD15} and HED~\cite{Ref:Xie15}. The whole network is fine-tuned from an initialization with the pre-trained VGG 16-layer net~\cite{Ref:Simonyan14}.
\paragraph{Model parameters} The hyper parameters of our network include: mini-batch size(10), base learning rate ($1\times10^{-6}$), loss weight for each side-output (1), momentum (0.9), initialization of the nested filters(0), initialization of of the scale-specific weighted fusion layer ($1/n$, where $n$ is the number of sliced scale-specific map), the learning rate of the scale-specific weighted fusion layer ($5\times10^{-6}$), weight decay ($2\times10^{-4}$), maximum number of training iterations ($20,000$).
\paragraph{Data augmentation}
Data augmentation is a principal way to generate sufficient training data for learning a ``good'' deep network. We rotate the images to 4 different angles ($0^\circ$, $90^\circ$, $180^\circ$, $270^\circ$) and flip with different axis(up-down,left-right,no flip), then resize images to 3 different scales ($0.8$, $1.0$, $1.2$), totally leading to an augmentation factor of 36. Note that when resizing a groundtruth skeleton map, the scales of the skeleton pixels in it should be multiplied by a resize factor accordingly.
\subsection{Performance Comparison}
We conduct our experiments by comparing our method FSDS with many others, including a traditional image processing method (Lindeberg's method~\cite{Ref:Lindeberg98}), three learning based segment linking methods ( Levinshtein's method~\cite{Ref:Levinshtein09}, Lee's method~\cite{Ref:Lee13} and Particle Filter~\cite{Ref:Widynski14}), three per-pixel classification/regression methods (Distance Regression~\cite{Ref:SironiLF14}, MIL~\cite{Ref:Tsogkas12} and MISL~\cite{Ref:Shen16}) and a deep learning based method (HED~\cite{Ref:Xie15}). For all theses methods, we use the source code provided by authors under default setting. For HED and FSDS, we perform iterations of sufficient numbers to obtain optimal models, $15,000$ and $18,000$ iterations for FSDS and HED, respectively. We apply a standard non-maximal suppression algorithm~\cite{Ref:DollarZ15} to the response maps of HED and ours to obtain the thinned skeletons for performance evaluation.
\subsubsection{Evaluation Protocol}
We follow the evaluation protocol used in~\cite{Ref:Tsogkas12}, under which the performances of skeleton extraction methods are measured by their maximum F-meansure ($\frac{2\cdot\text{Precision}\cdot\text{Recall}}{\text{Precision}+\text{Recall}}$) as well as precision-recall curves with respect to the groundtruth skeleton map. To obtain the precision-recall curves, the detected symmetry response is first thresholded into a binary map, which is then matched with the groundtruth skeleton map. The matching allows small localization errors
between detected positives and groundtruths. If a detected positive is matched with at least one groundtruth skeleton pixel, it is classified as true positive. In contrast, pixels that do not correspond to any groundtruth skeleton pixel are false positives. By assigning different thresholds to the detected skeleton response, we can obtain a sequence of precision and recall pair, which is used
to plot the precision-recall curve.
\subsubsection{SK506}
\begin{figure}[!h]
\centering
\includegraphics[width=0.8\linewidth]{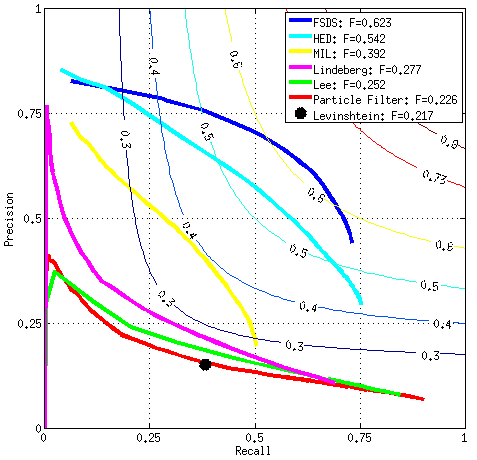}
\caption{Evaluation of skeleton extractors on the SK506 dataset. Leading skeleton extraction methods are ranked according to their best F-measure with respect to groundtruth skeletons. Our method, FSDS achieves the top result and shows both improved recall and precision at most of the
precision-recall regime. See Table~\ref{tbl:sk506} for more details about the other two quantities and method citations.\label{fig:pr_sk}
}\end{figure}

\begin{figure}[!h]
\centering
\includegraphics[width=1.0\linewidth]{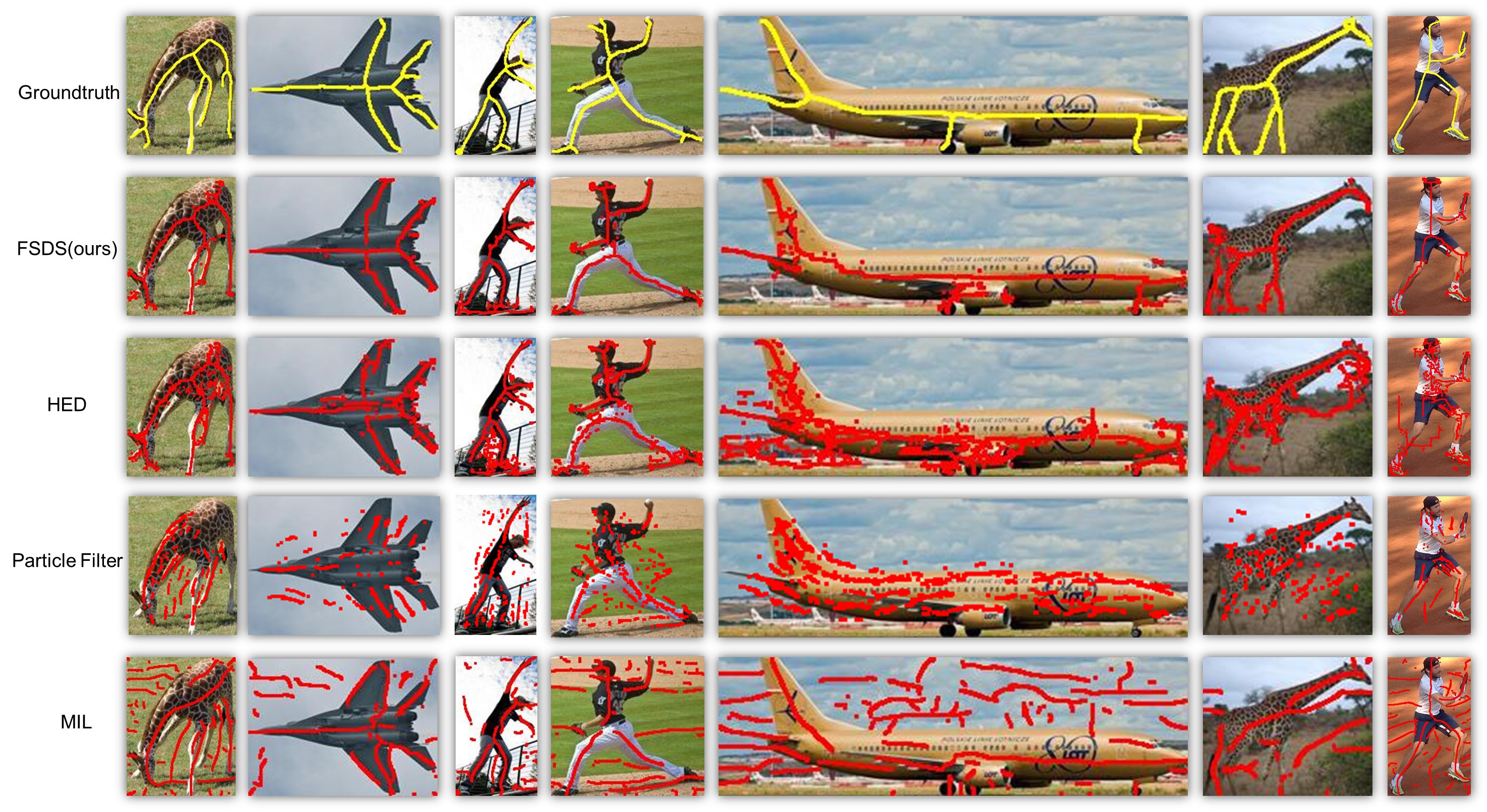}
\caption{Illustration of skeleton extraction results on the SK506 dataset for several selected images. The groundtruth skeletons are in yellow and the thresholded extraction results are in red. Thresholds were optimized over the whole dataset.\label{fig:examples_sk}
}\end{figure}

We first conduct our experiments on our newly built SK506 Dataset. Object skeletons in this dataset have large variances in both structures and scales. We split this dataset into 300 training and 206 testing images. We report the F-measure as well as the average runtime per image of each method on this dataset in Table.~\ref{tbl:sk506}. Observed that, both traditional image processing and per-pixel/segment learning methods perform not well, indicating the difficulty of this task. In addition, the segment linking methods are extremely time consuming. Our method FSDS outperforms others significantly, even compared with the deep learning based method HED. Besides, thanks to the powerful convolution computation ability of GPU, our method can process images in real time, about 20 images per second. The precision/recall curves shown in Fig.~\ref{fig:pr_sk} evidence again that FSDS is better than the competitors, as ours shows both improved recall and precision at most of the precision-recall regimes. We illustrate the skeleton extraction results obtained by several methods in Fig.~\ref{fig:examples_sk} for qualitative comparison. These qualitative examples show that our method hits on more groundtruth skeleton points and meanwhile suppresses the false positives. The false positives in the results of HED are probably introduced across response of different scales. Benefited from scale-associated learning and scale-specific fusion, our method is able to suppress such false positives.

\begin{table}[!h]
\centering
\caption{Performance comparison between different methods on the SK506 dataset. $\dag$GPU time.}\label{tbl:sk506}
\begin{tabular}{ccc}
\toprule
Method&F-measure&Avg Runtime (sec)\\
\midrule
Lindeberg~\cite{Ref:Lindeberg98}&0.227&4.03\\
Levinshtein~\cite{Ref:Levinshtein09}&0.218&144.77\\
Lee~\cite{Ref:Lee13}&0.252&606.30\\
Particle Filter~\cite{Ref:Widynski14}&0.226&322.25$\dag$\\
MIL~\cite{Ref:Tsogkas12}&0.392&42.38\\
HED~\cite{Ref:Xie15}&0.542&0.05$\dag$\\
\textbf{FSDS (ours)}&\textbf{0.623}&\textbf{0.05}$\dag$\\
\bottomrule
\end{tabular}
\end{table}

\subsubsection{WH-SYMMAX}
\begin{figure}[!h]
\centering
\includegraphics[width=0.8\linewidth]{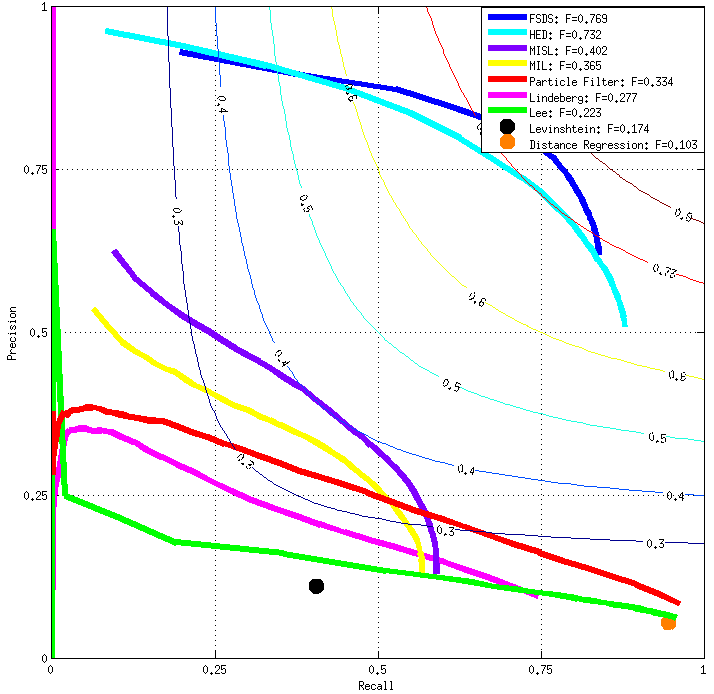}
\caption{Evaluation of skeleton extractors on the WH-SYMMAX Dataset~\cite{Ref:Shen16}. Leading skeleton extraction methods are ranked according to their best F-measure with respect to groundtruth skeletons. Our method, FSDS achieves the top result and shows both improved recall and precision at most of the
precision-recall regime. See Table~\ref{tbl:horse} for more details about the other two quantities and method citations.\label{fig:pr_horse}
}\end{figure}

\begin{figure}[!h]
\centering
\includegraphics[width=1.0\linewidth]{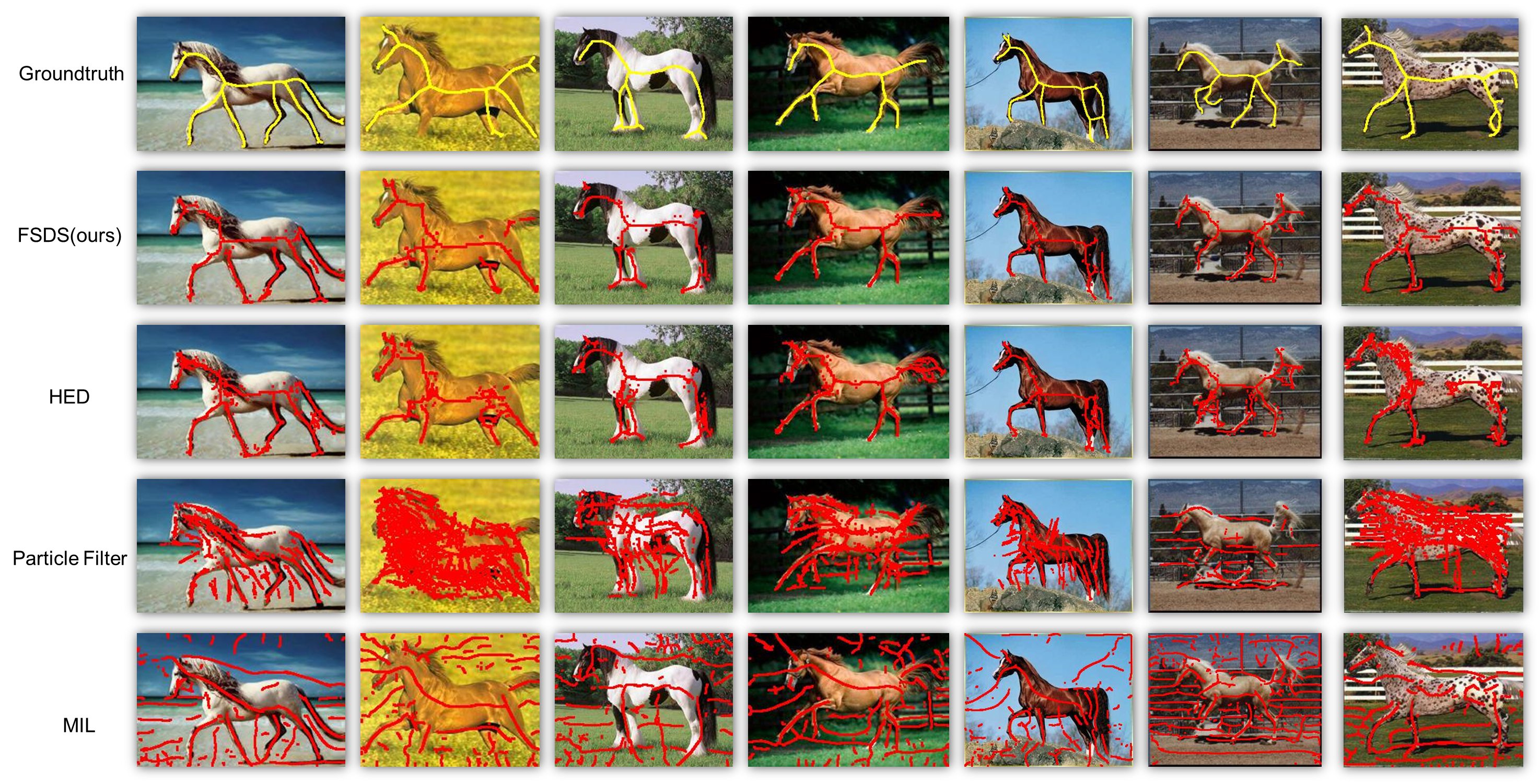}
\caption{Illustration of skeleton extraction results on the WH-SYMMAX dataset~\cite{Ref:Shen16} for several selected images. The groundtruth skeletons are in yellow and the thresholded extraction results are in red. Thresholds were optimized over the whole dataset.\label{fig:examples_horse}
}\end{figure}

The WH-SYMMAX dataset~\cite{Ref:Shen16} totally contains 328 images, among which the first 228 are used for training and the rest are used for testing. The precision/recall curves of skeleton extraction methods are shown in Fig.~\ref{fig:pr_horse} and summary statistics are in Table~\ref{tbl:horse}. Qualitative comparisons are
illustrated in Fig.~\ref{fig:examples_horse}. Both quantitative and qualitative results demonstrate that our method is clearly better than others.
\begin{table}[!h]
\centering
\caption{Performance comparison between different methods on the WH-SYMMAX Dataset~\cite{Ref:Shen16}. $\dag$GPU time.}\label{tbl:horse}
\begin{tabular}{ccc}
\toprule
Method&F-measure&Avg Runtime (sec)\\
\midrule
Lindeberg~\cite{Ref:Lindeberg98}&0.277&5.75\\
Levinshtein~\cite{Ref:Levinshtein09}&0.174&105.51\\
Lee~\cite{Ref:Lee13}&0.223&716.18\\
Particle Filter~\cite{Ref:Widynski14}&0.334&13.9$\dag$\\
Distance Regression~\cite{Ref:SironiLF14}&0.103&5.78\\
MIL~\cite{Ref:Tsogkas12}&0.365&51.19\\
MISL~\cite{Ref:Shen16}&0.402&78.41\\
HED~\cite{Ref:Xie15}&0.732&0.06$\dag$\\
\textbf{FSDS (ours)}&\textbf{0.769}&0.07$\dag$\\
\bottomrule
\end{tabular}
\end{table}

\subsubsection{SYMMAX300}
\begin{figure*}[!t]
\centering
\includegraphics[width=1.0\linewidth]{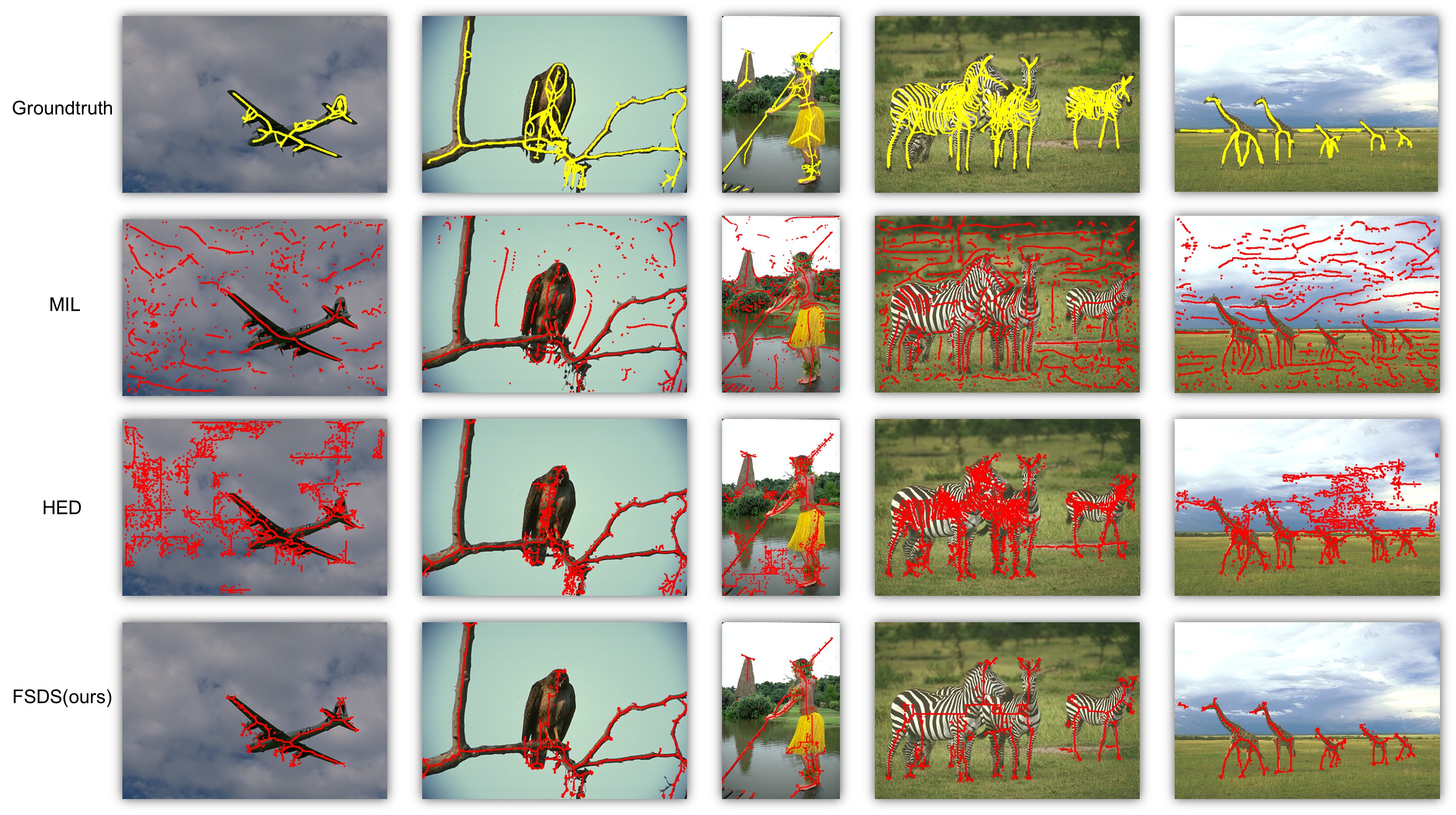}
\caption{Illustration of skeleton extraction results on the SYMMAX300 dataset~\cite{Ref:Tsogkas12} for several selected images. The groundtruth skeletons are in yellow and the thresholded extraction results are in red. Thresholds were optimized over the whole dataset.\label{fig:examples_sym}
}\end{figure*}

As we discussed in Sec.~\ref{sec:intro}, a large number of groundtruths are labeled on non-object parts in SYMMAX300~\cite{Ref:Tsogkas12}, which do not have organized structures as object skeletons. Our aim is to suppress those on non-object parts, so that the obtained skeletons can be used for other potential applications. In addition, the groundtruths for scale are not provided by SYMMAX300. Therefore, we do not evaluate our method quantitatively on SYMMAX300. Even so, Fig.~\ref{fig:examples_sym} shows that our method can obtain good skeletons of some objects in SYMMAX300. We also observe that the results obtained by our method have significantly less noises on background.

\subsubsection{Cross Dataset Generalization}
One may concern the scale-associated side outputs learned from one dataset might lead to higher generalization error when applying them to another dataset. To explore whether this is
the case, we test the model learned from one dataset on another one. For comparison, we list the cross dataset generalization results of MIL~\cite{Ref:Tsogkas12}, HED~\cite{Ref:Xie15} and our method in Table~\ref{tbl:cross}. Our method achieves better cross dataset generalization results than both the ``non-deep'' method (MIL) and the ``deep'' method (HED).

\begin{table}[!htbp]
\centering
\caption{Cross-dataset generalization results. TRAIN/TEST indicates the training/testing dataset used.}\label{tbl:cross}
\begin{tabular}{ccc}
\toprule
Method&Train/Test&F-measure\\
\midrule
MIL~\cite{Ref:Tsogkas12}&SK506/WH-SYMMAX&0.363\\
HED~\cite{Ref:Xie15}&SK506/WH-SYMMAX&0.637\\
\textbf{FSDS (ours)}&SK506/WH-SYMMAX&\textbf{0.692}\\
\hline
MIL~\cite{Ref:Tsogkas12}&WH-SYMMAX/SK506&0.387\\
HED~\cite{Ref:Xie15}&WH-SYMMAX/SK506&0.492\\
\textbf{FSDS (ours)}&WH-SYMMAX/SK506&\textbf{0.529}\\

\bottomrule
\end{tabular}
\end{table}

\subsubsection{Symmetric Part Segmentation} \label{sec:part_seg}
\begin{figure}[!t]
\centering
\includegraphics[width=0.8\linewidth]{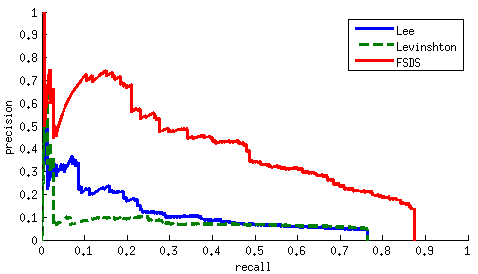}
\caption{Symmetric part segmentation results on BSDS-Parts dataset~\cite{Ref:Lee13}.\label{fig:part_detection}
}\end{figure}

\begin{figure}[!h]
\centering
\includegraphics[width=1.0\linewidth]{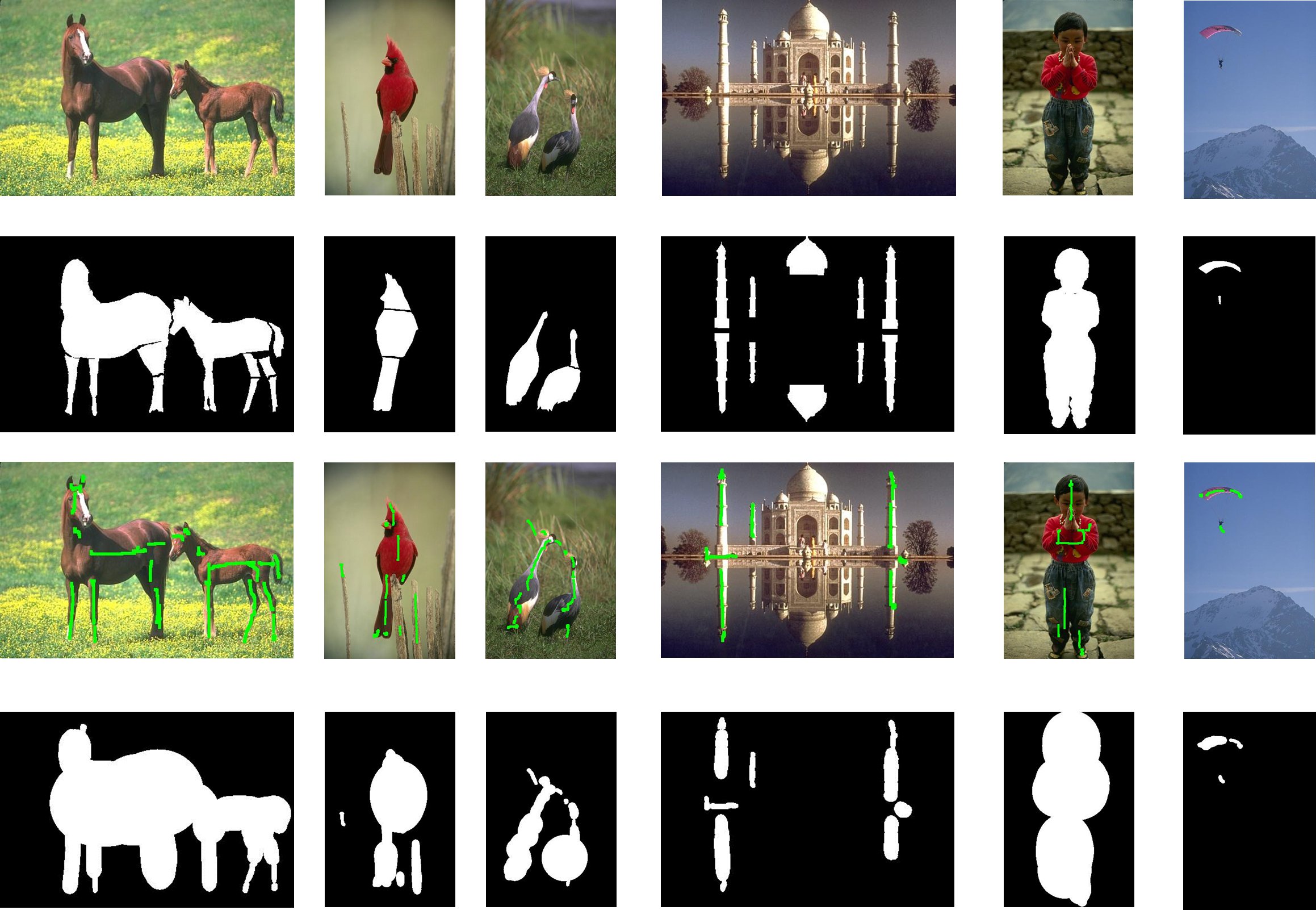}
\caption{Illustration of symmetric part segmentation results on the BSDS-Parts dataset~\cite{Ref:Lee13} for several selected images. In each column, we show the orginal image, the segmentation groundtruth, the thresholded extracted skeleton (in green), the segmented masks recovered by the skeleton. Thresholds were optimized over the whole dataset.\label{fig:examples_bsds}
}\end{figure}

Part-based models are widely used for object detection and recognition in natural images~\cite{Ref:Felzenszwalb10,Ref:Zhu15}. To verify the usefulness of the extracted skeletons, we follow the criteria in~\cite{Ref:Levinshtein09} for symmetric part segmentation. We evaluate the ability of our skeleton to find segmentation masks corresponding to object parts in a cluttered scene. Our network provides a predicted scale for each skeleton pixel (the fused skeleton score map for which scale has maximal response). With it we can recover object parts from skeletons. For each skeleton pixel $x_j$, we can predict its scale by $\hat{s}_j=\sum_{i=1}^M\text{Pr}(z_j=i|X;\mathbf{\Theta}\ast,\mathbf{\Phi}\ast,\mathbf{a}_0\ast)r_i$. Then for a skeleton segment $\{x_j,j=1,\ldots,N\}$, where $N$ is the number of the skeleton pixels in this segment, we can obtain a segmented object part mask by $\mathcal {M}=\bigcup_{j=1}^ND_j$, where $D_j$ is the disk of center $x_j$ and diameter $\hat{s}_j$. A confidence score is also assigned to each object part mask for quantitative evaluation: $P_{\mathcal {M}}=\frac{1}{N}\sum_{j=1}^N(1-\text{Pr}(z_j=0|X;\mathbf{\Theta}\ast,\mathbf{\Phi}\ast,\mathbf{a}_0\ast))$. We compare our segmented part masks with Lee's method~\cite{Ref:Lee13} and Levinshtein's method~\cite{Ref:Levinshtein09} on their BSDS-Parts dataset~\cite{Ref:Lee13}, which contains 36 images annotated with ground-truth masks corresponding to the symmetric parts of prominent objects. The segmentation results are evaluated by the protocol used in~\cite{Ref:Lee13}:  A segmentation mask $\mathcal {M}_{seg}$ is counted as a hit if its overlap with the ground-truth mask $\mathcal {M}_{gt}$ is greater than $0.4$, where overlap is measured by intersection-over-union (IoU). A precision/recall curve is obtained by varying a threshold over the confidence scores of segmented masks. The quantitative evaluation results are summarized in Fig.~\ref{fig:part_detection}, which indicate a significant improvement over the other two methods. Some qualitative results on the BSDS-Parts dataset~\cite{Ref:Lee13} are shown in Fig.~\ref{fig:examples_bsds}.

\subsubsection{Object Proposal Detection} \label{sec:detection}
\begin{figure}[!h]
\centering
\includegraphics[width=1.0\linewidth]{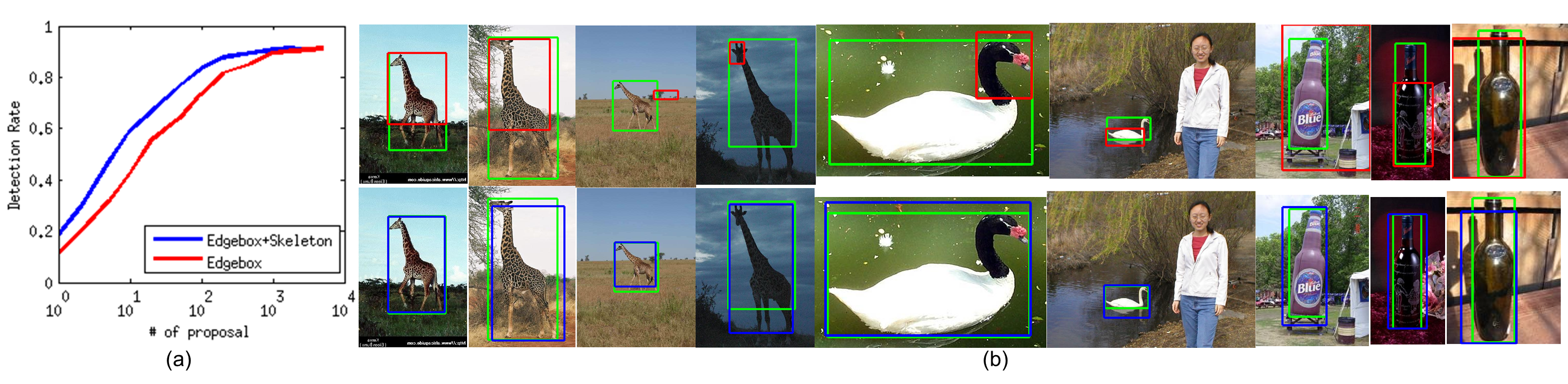}
\caption{Object proposal results on ETHZ Shape Classes~\cite{Ref:Ferrari06}. (a) The curve (IoU = 0.7). (b) Examples. Groundtruth (green), the closest proposal to groundtruth of Edgebox (red) and ours (blue).
}\label{fig:object_proposals}
\end{figure}

To demonstrate the potential of the extracted skeletons in object detection, we do an experiment on object proposal detection. Let $h_B^E$ be the objectness score of a bounding box $B$ obtained by Edge Boxes~\cite{Ref:ZitnickD14}, we define our objectness score by $h_B=\frac{\sum_{\forall\mathcal {M}{\cap}B\neq\emptyset}\mathcal {M}{\cap}B}{\sum_{\forall\mathcal {M}{\cap}B\neq\emptyset}\mathcal {M}+\epsilon}{\cdot}h_B^E$, where $\epsilon$ is a very small number to ensure the denominator to be non-zero, and $\mathcal {M}$ is a part mask reconstructed by a detected skeleton segment. As shown in Fig.~\ref{fig:object_proposals}, the new scoring method achieves a better object proposal detection result than Edge Boxes.

\section{Conclusion}
We have presented a fully convolutional network with multiple scale-associated side outputs to extract skeletons from natural images. By pointing out the relationship between the receptive field sizes of the sequential stages in the network and the skeleton scales they can capture, we emphasized the important roles of the proposed scale-associated side outputs in (1) guiding multi-scale feature learning and (2) fusing scale-specific responses from different stages. The encouraging experimental results demonstrate the effectiveness of the proposed method for skeleton extraction from natural images. It achieves significant improvements to all other competitors.

\noindent {\bf Acknowledgement}. This work was supported in part by
the National Natural Science Foundation of China under Grant
61303095 and 61573160, in part by Research Fund for the Doctoral
Program of Higher Education of China under Grant 20133108120017, and in
part by Innovation Program of Shanghai Municipal Education
Commission under Grant 14YZ018. We thank NVIDIA Corporation
for providing their GPU device for our academic research.
{\small
\bibliographystyle{ieee}
\bibliography{egbib}
}

\end{document}